\documentclass[letterpaper, 10 pt, journal, twoside]{ieeetran}

\usepackage{graphicx}
\usepackage{xpatch, amsmath,amssymb,bm,mathtools}
\usepackage[capitalize]{cleveref}
\usepackage{xcolor}
\usepackage{fp, tikz, pgfplots, multirow, nicefrac}
\usepackage{siunitx}
\usepackage{subcaption}
\usetikzlibrary{arrows}
\usetikzlibrary{shapes}
\usetikzlibrary{patterns}
\usetikzlibrary{decorations.pathmorphing}
\usetikzlibrary{decorations.pathreplacing}
\usetikzlibrary{decorations.shapes}
\usetikzlibrary{decorations.text}
\usetikzlibrary{shapes.misc}
\usetikzlibrary{decorations.markings}
\usetikzlibrary{arrows.meta}
\usetikzlibrary{backgrounds}
\usepgfplotslibrary{external}
\usepgfplotslibrary{fillbetween}
\usepgfplotslibrary{statistics}
\usepackage[norelsize,ruled,linesnumbered,noend]{algorithm2e}
\usepackage{setspace}
\usepackage{caption} 
\newtheorem{definition}{Definition}
\newtheorem{problem}{Problem}
\newtheorem{remark}{Remark}

\newtheorem{assumption}{Assumption}

\newcommand{\defeq}{\vcentcolon=}

\newcommand{\R}{\mathbb{R}}
\newcommand{\tp}{^\mathsf{T}}
\newcommand{\diff}{\mathop{}\!\mathrm{d}}

\newlength{\fwidth}
\newlength{\fheight}

\IEEEoverridecommandlockouts                              

\title{
Vision-Based Uncertainty-Aware Motion Planning based on Probabilistic Semantic Segmentation
}

\author{Ralf R\"omer$^*$, Armin Lederer$^*$, Samuel Tesfazgi, and Sandra Hirche%
\thanks{Manuscript received: April, 29, 2023; Revised July, 26, 2023; Accepted September, 26, 2023.}
\thanks{This paper was recommended for publication by Editor J. Kober upon evaluation of the Associate Editor and Reviewers' comments.
This work was supported by the European Research Council (ERC)
Consolidator Grant ”Safe data-driven control for human-centric systems
(CO-MAN)” under grant agreement number 864686, by the Horizon
2020 research and innovation programme of the European Union under
grant agreement number 871767 of the project ReHyb, and by TUM AGENDA 2030, funded by the Federal Ministry of Education and Research (BMBF) and the Free State of Bavaria under the Excellence Strategy of the Federal Government and the L\"ander as well as by the Hightech Agenda Bavaria.} 
\thanks{The authors are with the TUM School of Computation, Information and Technology, Technical University of Munich, 80333 Munich, Germany. Ralf Römer is with the Learning Systems and Robotics Lab (LSY). Armin Lederer, Samuel Tesfazgi, and Sandra Hirche are with the Chair of Information-Oriented Control (ITR).
{\tt\footnotesize \{ralf.roemer; armin.lederer; samuel.tesfazgi; hirche\}@tum.de}}
\thanks{$^*$ Both authors contributed equally.}
\thanks{Digital Object Identifier (DOI): see top of this page.}
}

\begin{document}

\maketitle


\begin{abstract}
    For safe operation, a robot must be able to avoid collisions in uncertain environments. 
    Existing approaches for motion planning under uncertainties often assume parametric obstacle representations and Gaussian uncertainty, which can be inaccurate. 
    While visual perception can deliver a more accurate representation of the environment, its use for safe motion planning is limited by the inherent miscalibration of neural networks and the challenge of obtaining adequate datasets. 
    To address these limitations, we propose to employ ensembles of deep semantic segmentation networks trained with massively augmented datasets to ensure reliable probabilistic occupancy information. To avoid conservatism during motion planning, we directly employ the probabilistic perception in a scenario-based path planning approach.
    A velocity scheduling scheme is applied to the path
    to ensure a safe motion despite tracking inaccuracies.
    We demonstrate the effectiveness of the massive data augmentation in combination with deep ensembles and the proposed scenario-based planning approach in comparisons to state-of-the-art methods and validate our framework in an experiment with a human hand as an obstacle.
\end{abstract}

\begin{IEEEkeywords}
Planning under Uncertainty, Object Detection, Segmentation and Categorization, Deep Learning for Visual Perception.
\end{IEEEkeywords}

\section{Introduction}
\IEEEPARstart{W}{hen} robots operate in the real world, 
safety requires the avoidance of unintended collisions. In order to detect any form of obstacles, visual perception using deep learning (DL) has gained growing attention in recent years
. While DL-based perception systems have achieved impressive results in various tasks, 
several issues prevent their applicability in many safety-critical systems. Firstly, DL models are typically trained using large datasets \cite{Sun17, Shelhamer17},
which are often not available for custom robotics tasks. However, the creation and annotation of task specific datasets is generally costly, such that small datasets are desirable. Moreover, DL models are prone to causing prediction errors in previously unseen situations~\cite{Kendall17}, 
such that a reliable quantification of their predictive uncertainty is important to enable a cautious behavior of the robotic system. Finally, even when uncertainty information is available, it is often not in the parametric form required for uncertainty-aware planning algorithms, such that additional simplifications are necessary \cite{Zhu2019}.\looseness=-1 

\subsection{Related Work}
A common approach in existing work is to address visual perception by semantic segmentation \cite{Milioto2019}. 
In recent years, the focus of segmentation approaches has been on DL architectures due to the impressive accuracy achieved with fully convolutional networks \cite{Shelhamer17}. 
%
However, DL is inherently miscalibrated and therefore often produces over-confident predictions \cite{Kendall17}, 
such that research has focused on improving uncertainty quantification of DL. 
While Bayesian neural networks with dropout have attracted interest early on \cite{Gal16}, deep ensembles \cite{Lakshminarayanan17}
have gained increasing attention 
due to their often demonstrated superior performance, 
reasonable computational cost for inference and parallelizability \cite{ Ovadia2019}
.
These advantages have also been shown when applying them to semantic segmentation~\cite{Mehrtash2020},
but they typically require large training datasets and do not extend to low data regimes, in which ensembling alone is generally insufficient to obtain well-calibrated uncertainty estimates~\cite{rahaman2021}. 
In order to mitigate this limitation, 
numerous methods for data augmentation have been proposed~\cite{Devries2017,Zhang2018} with the goal of improving the generalizability and robustness.
The main focus of these augmentation schemes is usually the avoidance of overfitting, such that the dataset size is commonly increased by merely a low factor, often two, and by applying not more than three augmentation methods~\cite{Uzun2021}. 
However, this is insufficient to achieve robust calibration for small task-specific datasets which are often not representative in terms of diversity.\looseness=-1

The results of semantic segmentation can directly be used for robotic motion planning, but such approaches ignore uncertainties in the perception \cite{Bartolomei2020}. 
Previous works on motion planning with probabilistic environment representation mostly assume known obstacle geometry and Gaussian distributed object position \cite{Zhu2019, Park2018, Kamel2017}. 
While these assumptions allow for the derivation of analytical chance constraints, a coarse over-approximation of obstacles can lead to excessive conservatism of planned paths in complex environments. A finer obstacle parameterization or the individual representation of each obstacle can become computationally challenging, especially when the number of obstacles is large, and representing a variation of the uncertainty in the vicinity of the obstacles is difficult. 
Safe perception-based navigation has also been addressed using Hamilton-Jacobi reachability \cite{Bajcsy.2019}, but the computational cost increases significantly for irregular obstacle surfaces.
Other approaches rely on deterministic error bounds for positions estimated using visual perception, such that robust planning methods can be employed \cite{Dean2020}.
Since these error bounds require dense training data in practice, large datasets must generally be available for such approaches.
Hence, there exists no uncertainty-aware and flexible approach for safe motion planning based on semantic segmentation when merely a small, non-representative dataset is available.

\subsection{Contribution}
We propose a general framework for motion planning with uncertainty based on visual perception via probabilistic semantic segmentation.
In the perception module, we enable the training of well-calibrated semantic segmentation models for small, non-representative datasets by combining deep ensembles with massive data augmentation.
By modifying training images with eight different methods, we increase the dataset size by a factor of 20, which allows us to obtain reliable probabilistic occupancy information.
In the safe motion planning module, we avoid conservatism and computational complexity due to parametric obstacle representations by formulating the path planning problem in an uncertain environment as a scenario optimization problem. This allows us to directly determine collision probabilities using the results from  probabilistic semantic segmentation, which we exploit in a scenario chance constrained version of the popular RRT* algorithm \cite{Karaman2011}.
For the resulting path, we propose safe velocity scheduling to ensure safe robot motion despite tracking inaccuracies.
We demonstrate the effectiveness of the proposed data augmentation approach for visual perception and the scenario-based path planning formulation in a comparison to state-of-the-art methods. Moreover, we evaluate the safety and robustness of our framework in an experiment with a robotic manipulator.\looseness=-1
\section{Problem Statement}
We consider widely-used camera-based perception based on a DL model, which requires labeled data for training. However, due to the lack of suitable training data for many specific robotic applications and the inherent miscalibration of neural networks \cite{Kendall17}, it is generally difficult to obtain an accurate representation of the environment with reliable information on uncertainty.
Therefore, we investigate this problem in this letter as formalized in the following.


\begin{problem}[Uncertainty in visual perception]
\label{prob1}
    Assume a small dataset with low variety is given, which contains RGB images $\bm{C}\!\in\!\mathbb{R}^{H\times W\times 3}$ with height/width $H$/$W$ of marked obstacles with arbitrary pose, geometry and size. Based on the dataset, 
    we consider the problem of training a DL model $\bm{f}:\mathbb{R}^{H\times W\times 3}\times \Theta\rightarrow [0,1]^{H\times W}$ with parameters $\bm{\theta}\!\in\!\Theta$ which outputs for each point $(i,j)$ in an image $\bm{C}$ the probability of being occupied by an obstacle $\mathcal{O}$, i.e., $P((i,j)\!\in\!\mathcal{O})\!=\!f_{ij}(\bm{C},\bm{\theta})$.\looseness=-1
\end{problem}

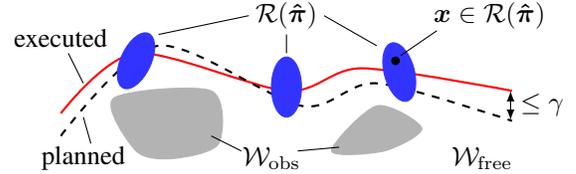
\begin{figure}
\centering

\begin{tikzpicture}
    \draw [red, thick] plot [smooth] coordinates {(0,0.2) (1,1) (3,0.5) (4,0.8) (6,0.5)};
    \draw [black, thick, dashed] plot [smooth] coordinates {(0,-0.1) (1.3,1.1) (3.2,0.3) (4.2,0.6) (6,0.1)};
    
    \draw [fill=gray!60, gray!60] plot [smooth cycle] coordinates {(1,-0.3) (0.7,0.4) (2,0.5) (2.1,-0.2) (1.5,-0.4)};
    
    \draw [fill=gray!60, gray!60] plot [smooth cycle] coordinates {(3.6,-0.3) (4.2,0.3) (4.8,0) (4.3,-0.3)};

    \draw[rotate around={-25:(1,0.9)},blue!80, fill=blue!80] (1,0.9) ellipse (0.2 and 0.4);
    \draw[blue!80, fill=blue!80] (3,0.55) ellipse (0.2 and 0.4);
    \draw[rotate around={15:(4.5, 0.75)},blue!80, fill=blue!80] (4.5, 0.75) ellipse (0.2 and 0.4);

    \draw[latex-latex] (5.98,0.5) to[] (5.98,0.11);
    \node at (6.4,0.31) {$\leq \gamma$};

    \node at (2.8,-0.4) {$\mathcal{W}_\mathrm{obs}$};
    \draw (3.15,-0.3) -- (3.7,-0.25);
    \draw (2.4,-0.3) -- (2,-0.1);
    \node at (5.6,-0.4) {$\mathcal{W}_\mathrm{free}$};
    \node at (3,1.5) {$\mathcal{R}(\bm{\hat{\pi}})$};
    \draw (3.5,1.5) -- (4.2,1.1);
    \draw (3,1.25) -- (3,1);
    \draw (2.5,1.5) -- (1.3,1.25);
    
    \node at (0,1.2) {executed};
    \draw (0.1,1) -- (0.3,0.6);
    \node at (0.3,-0.4) {planned};
    \draw (0.4,-0.2) -- (0.3,0.15);
    \draw[black, fill=black] (4.45,0.9) circle(0.05);
    \node at (5.7,1.5) {$\bm{x}\in \mathcal{R}(\bm{\hat{\pi}})$};
    \draw (5,1.3) -- (4.52,0.97);
\end{tikzpicture}

\caption{We aim to plan a trajectory $\bm{\pi}:[0,\tau]\rightarrow \mathcal{T}$ in the task space such that during the executed motion $\bm{\hat{\pi}}(\cdot)$, any point $\bm{x}\in\mathcal{R}(\bm{\hat{\pi}}(t))$ occupied by the robot $\mathcal{R}(\bm{\hat{\pi}}(t))$ lies inside the free workspace $\mathcal{W}_\mathrm{free}$ with a probability of at least $1-\delta$ for all $t\in[0,\tau]$.}
\label{fig:planning_idea}    
\end{figure}

Based on the probabilities of obstacles in the image space, the robotic system must be capable of planning a safe trajectory of poses {$\{\bm{p}|\,\bm{p}\in\mathcal{T}\}$} in the task space $\mathcal{T}$. Since each pose $\bm{p}$ implies that the robot occupies some region $\mathcal{R}(\bm{p})\subset\mathcal{W}$ of the physical workspace $\mathcal{W} \subseteq \R^3$, a pose $\bm{p}$ is only collision-free if the set $\mathcal{R}(\bm{p})$ lies completely in the obstacle-free subset $\mathcal{W}_{\mathrm{free}}\subset\mathcal{W}$. 
However, this cannot be ensured deterministically in general as merely an uncertain estimate of the obstacles and consequently $\mathcal{W}_{\mathrm{free}}$ is available from the visual perception module. 
Therefore, we aim to satisfy the constraints imposed by $\mathcal{W}_\mathrm{free}$ probabilistically via individual chance constraints with a prescribed probability threshold $\delta\in(0,1)$.
This leads to the following formal definition of a $\delta$-safe motion, which is illustrated in \cref{fig:planning_idea}.
\begin{definition}[$\delta$-safe motion]
    The motion $\hat{\bm{\pi}}:[0,\tau]\rightarrow \mathcal{T}$ executed within the time interval $[0,\tau]$ is called $\delta$-safe if it satisfies $P(\bm{x} \in \mathcal{W}_\mathrm{free}) \geq 1-\delta$, ${\forall \bm{x} \in \mathcal{R}(\hat{\bm{\pi}}(t))}$, $\forall t \in [0,\tau]$.
\end{definition}

In this definition, safety is introduced using an individual condition for each time instance, which is a commonly considered requirement for planning in uncertain environments~\cite{Zhu2019, Park2018}. 
However, these conditions are posed on the motion $\hat{\bm{\pi}}(\cdot)$ \textit{realized by the robotic system}, which is a significantly stronger notion of safety than merely requiring their satisfaction for the \textit{planned trajectory} $\bm{\pi}:[0,\tau]\rightarrow \mathcal{T}$, which is also illustrated in Fig. \ref{fig:planning_idea}. Therefore, this safety notion clearly cannot be ensured without additional information about the dynamics of the robotic system, such that we assume the availability of a control law with guaranteed tracking error bounds as formalized in the following.
\begin{assumption}[Velocity-dependent tracking error bound]\label{ass:track_error}
    The tracking error of the robotic system is bounded by a non-decreasing function $\gamma:\R_0^+\rightarrow \R_0^+$
    of its reference velocity, i.e.,
     $\|\hat{\bm{\pi}}(t)-\bm{\pi}(t)\|_2\leq\gamma(\|\dot{\bm{\pi}}(t)\|)$.
\end{assumption}

This assumption reflects the fact that the tracking error of robotic systems often grows with increasing reference velocity due to unmodeled effects in control laws, e.g., friction or imprecise inertia parameters. 
Note that it does not require exact tracking for zero velocity, such that it can also be employed for underactuated robots, which can exhibit relatively large tracking errors at low velocities. Reducing the conservatism for such systems using more sophisticated controllers, e.g., based on barrier functions \cite{Dean2020},
is subject to future work. \looseness=-1
{Tracking error bounds can be obtained, e.g., statistically from experiments \cite{Nubert2020}. However, generating test trajectories that cover a sufficiently wide range of operating conditions can be challenging for higher-dimensional task spaces.}

In addition to safety, other criteria {often have to be considered during planning. While they can be represented using general cost functions in principle, we restrict ourselves to commonly found path integrals over immediate costs $c:\mathcal{T}\rightarrow R_{0,+}$, such that their velocity-independence can be exploited to employ computationally more efficient solution approaches \cite{Bobrow1985}.}
Therefore, we address the following safe vision-based planning problem.
\looseness=-1
\begin{problem}[Safe vision-based motion planning]\label{prob2}
    Given the uncertain estimate of the obstacles obtained from the visual perception system and the velocity-dependent tracking error bound in \cref{ass:track_error}, we consider the problem of finding a trajectory $\bm{\pi}(\cdot)$, which minimizes the cost $c$ along the path defined through $\bm{\pi}(\cdot)$ and ensures $\delta$-safety of the resulting executed motion $\hat{\bm{\pi}}(\cdot)$, i.e., 
    \label{eq:prob_statement}
    \begin{align}\label{eq:prob_cost}
        &\min\limits_{\bm{\pi}(\cdot)} \int_{0}^{\tau} c(\bm{\pi}(t))\|\dot{\bm{\pi}}(t)\|_2\mathrm{d}t \quad \text{such that  $\hat{\bm{\pi}}(\cdot)$ is $\delta$-safe.} 
    \end{align}
\end{problem}
\section{Visual Perception with Uncertainty Representation}

In order to address \cref{prob1}, it is necessary to probabilistically solve a classification problem for each pixel, which is commonly referred to as semantic segmentation. Therefore, we briefly introduce the fundamentals of training DL models for semantic segmentation in \cref{subsec:semseg}, before we show the extension to an ensemble of DL models to obtain occupancy probabilities for each pixel in \cref{subsec:ensemble}. 
Finally, we develop a data augmentation approach to achieve more robust uncertainty estimation in \cref{subsec:augment}.

\subsection{Deep Learning for Semantic Segmentation}\label{subsec:semseg}

For classifying each pixel in an image using semantic segmentation, we employ the commonly used approach of fully convolutional networks (FCNs) with an encoder-decoder structure \cite{Shelhamer17}. 
The encoder applies convolutional and pooling layers, capturing contextual information in a feature vector with downsampled spatial dimensions. In the decoder, the spatial dimensions are upsampled back to the input image size $H \times W$, allowing for pixel-wise dense predictions. 
We apply atrous convolutional layers 
with different rates in parallel (Atrous Spatial Pyramid Pooling), which allows to capture objects at multiple scales~\cite{Chen2018}. In the final layer of the model, we employ the softmax activation function for the output channel of every image pixel $(i,j)$.
Denoting the parameters of the DL model by $\hat{\bm{\theta}}$, this yields a function $\hat{f}_{ij}(\cdot,\hat{\bm{\theta}})$ for every pixel, which outputs a probability-like value for pixel $(i,j)$ not being occupied by an obstacle, i.e., $\hat{f}_{ij}(\bm{C},\hat{\bm{\theta}})\in [0,1]$.
For the training process, we employ the commonly used cross-entropy loss ${\mathcal{L}'(y_{ij},\!\hat{\bm{\theta}})\!=\!-y_{ij}\!\log\!\big(\hat{f}_{ij}(\bm{C},\!\hat{\bm{\theta}})\!\big)}\!-\!{(1-y_{ij})\!\log\!\big(1\!-\!\hat{f}_{ij}(\bm{C},\!\hat{\bm{\theta}})\!\big)}$, where $y_{ij} = 0$ if $(i,j)\in\mathcal{O}$, i.e., pixel $(i,j)$ lies in an obstacle, and $y_{ij}=1$ otherwise. It is computed pixel-wise and summed over the spatial dimensions of the final layer, resulting in the total loss $\mathcal{L}(\hat{\bm{\theta}}) = \sum_{ij} \mathcal{L}'(y_{ij},\hat{\bm{\theta}})$.\looseness=-1

\subsection{Probabilistic Segmentation using Deep Ensembles}\label{subsec:ensemble}
\begin{figure}
	\centering
\begin{tikzpicture}
    \node at (0,-0.5) {\includegraphics[width=2.4cm]{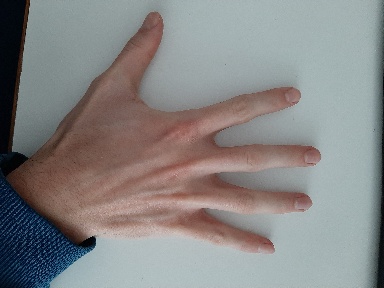}};
    \node at (0cm,0.6cm) {\small Original image};

    \node at (-3.3,-0.02) {\includegraphics[width=1.8cm]{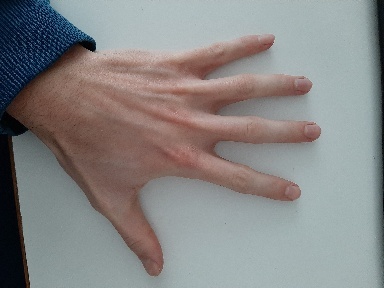}};
    \node[align=center,draw=black, rectangle,minimum height=1.7cm, minimum width=2cm, rounded corners, thick] at (-3.3cm,0.12cm) {};
    \node at (-3.3cm,0.8cm) {\scriptsize Flipping};
    \draw[->, >=latex,thick] (-1.25,0) to[] (-2.25,0);

    \node at (-3.3,-1.82) {\includegraphics[width=1.8cm]{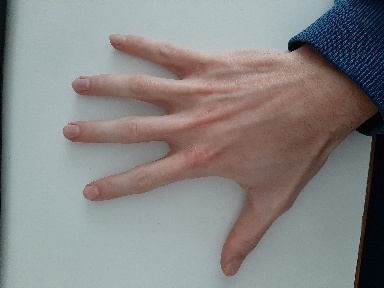}};
    \node[align=center,draw=black, rectangle,minimum height=1.7cm, minimum width=2cm, rounded corners, thick] at (-3.3cm,-1.68cm) {};
    \node at (-3.3cm,-1cm) {\scriptsize Rotation};
    \draw[->, >=latex,thick] (-1.25,-0.8) to[] (-2.25,-1.2);

    \node at (3.3,-0.02) {\includegraphics[width=1.8cm]{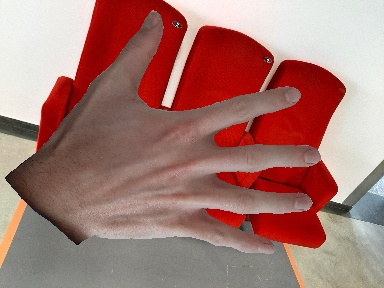}};
    \node[align=center,draw=black, rectangle,minimum height=1.7cm, minimum width=2cm, rounded corners, thick] at (3.3cm,0.12cm) {};
    \node at (3.3cm,0.8cm) {\scriptsize New background};
    \draw[->, >=latex,thick] (1.25,0) to[] (2.25,0);

    \node at (3.3,-1.82) {\includegraphics[width=1.8cm]{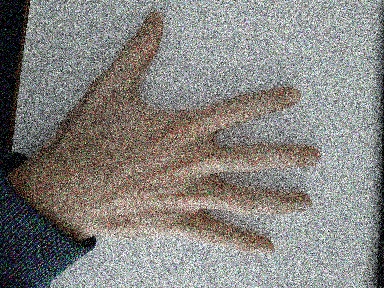}};
    \node[align=center,draw=black, rectangle,minimum height=1.7cm, minimum width=2cm, rounded corners, thick] at (3.3cm,-1.68cm) {};
    \node at (3.3cm,-1cm) {\scriptsize Noise};
    \draw[->, >=latex,thick] (1.25,-0.8) to[] (2.25,-1.2);

    \node at (-3.15,-3.62) {\includegraphics[width=1.8cm]{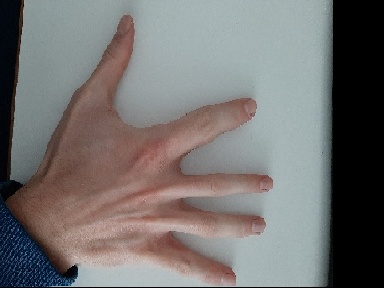}};
    \node[align=center,draw=black, rectangle,minimum height=1.7cm, minimum width=2cm, rounded corners, thick] at (-3.15cm,-3.48cm) {};
    \node at (-3.15cm,-2.8cm) {\scriptsize Grid distortion};
    \draw[->, >=latex,thick] (-1.1,-1.45) to[] (-2.2,-2.6);

    \node at (-1.05,-3.62) {\includegraphics[width=1.8cm]{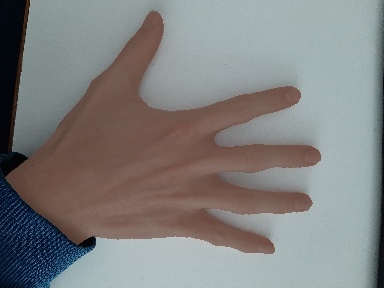}};
    \node[align=center,draw=black, rectangle,minimum height=1.7cm, minimum width=2cm, rounded corners, thick] at (-1.05cm,-3.48cm) {};
    \node at (-1.05cm,-2.8cm) {\scriptsize Color modification};
    \draw[->, >=latex,thick] (-0.4,-1.45) to[] (-0.8,-2.58);

    \node at (1.05,-3.62) {\includegraphics[width=1.8cm]{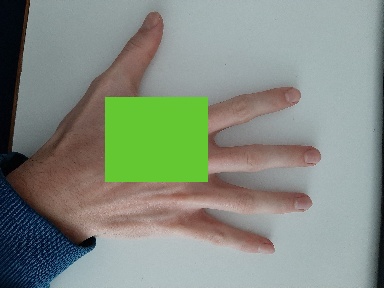}};
    \node[align=center,draw=black, rectangle,minimum height=1.7cm, minimum width=2cm, rounded corners, thick] at (1.05cm,-3.48cm) {};
    \node at (1.05cm,-2.8cm) {\scriptsize Random erase};
    \draw[->, >=latex,thick] (0.4,-1.45) to[] (0.8,-2.58);

    \node at (3.15,-3.62) {\includegraphics[width=1.8cm]{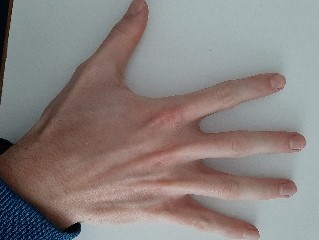}};
    \node[align=center,draw=black, rectangle,minimum height=1.7cm, minimum width=2cm, rounded corners, thick] at (3.15cm,-3.48cm) {};
    \node at (3.15cm,-2.8cm) {\scriptsize Cropping};
    \draw[->, >=latex,thick] (1.1,-1.45) to[] (2.2,-2.6);
\end{tikzpicture}

\caption{We employ eight methods for data augmentation to systematically add missing variety to the dataset. This increases robustness of the trained segmentation model to differences in perceived scenes and reduces the epistemic uncertainty typically caused by insufficient coverage of the input space by the training samples.}
\label{fig:augmentation_methods}
\end{figure}
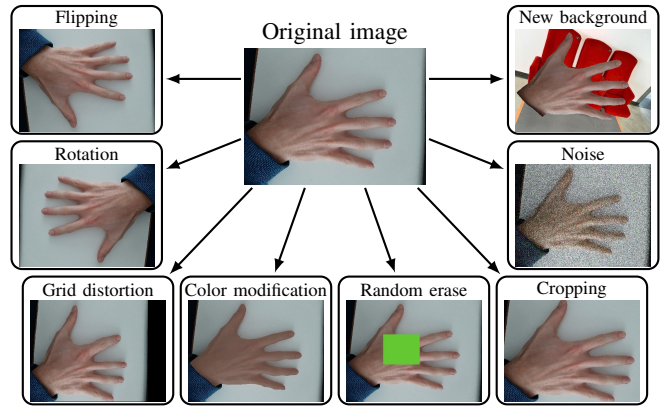

Even though DL models for semantic segmentation yield probability-like outputs, these values are generally not well suited as a measure of uncertainty due to the inherent miscalibration and overconfidence of NNs \cite{Kendall17}. 
We address this issue by employing an ensemble of models which is known to produce well calibrated uncertainty predictions with suitable training data \cite{Ovadia2019}. 
An ensemble consists of multiple distinct models, called ensemble members. For training the ensemble, we initialize each ensemble member with random model parameters and randomly shuffle the training data before each epoch \cite{Lakshminarayanan17}. The models are trained independently, such that they capture different features within the data. For inference, the individual member predictions are combined to the final prediction by considering the ensemble as a uniformly-weighted mixture of $M$ models, i.e., $\bm{f}(\bm{C},\bm{\theta})=\sum\nolimits_{m=1}^{M} \hat{\bm{f}}(\bm{C}, \hat{\bm{\theta}}_m)/M$,
where $\hat{\bm{\theta}}_m$ denotes the parameters of the $m$-th member concatenated into the overall parameter vector~$\bm{\theta}$. 
Since each function $f_{ij}(\cdot,\bm{\theta})$ also yields values in the range $[0,1]$, but generally exhibits a better calibration, 
we use it to determine the occupancy probabilities of pixels in the following.\looseness=-1

\subsection{Data Augmentation for Dataset Diversification}\label{subsec:augment}
\label{sec:meth_data_aug}

While deep ensembles can also be used for learning from small datasets in principle, it has been demonstrated that the resulting performance in semantic segmentation strongly depends on the size of the training set \cite{Sun17}. 
Moreover, due to the low variation in the training dataset considered in \cref{prob1}, the estimated epistemic uncertainty may not be well-calibrated when actually employing the deep ensemble in applications with more diverse images \cite{rahaman2021}.
Therefore, directly training a deep ensemble using a small dataset with low variation would be unreliable in safety-critical applications.\looseness=-1


In order to mitigate this effect, we propose to massively augment the training data to artificially add the missing variation to the data.
The underlying idea of this augmentation is to apply label-preserving transformations to the annotated images, as well as
to the corresponding segmentation masks~\cite{Shorten2019}. 
This idea can be exploited, e.g., when merely few annotated images of the different operating environments are available, to artificially modify their backgrounds as proposed in~\cite{Ghiasi2021}. Moreover, the robustness of the ensemble against partial object occlusions can be increased by randomly erasing parts of the labeled object \cite{zhong2020}. In total, we identify eight augmentation methods for straightforwardly adding missing information to segmentation datasets, which are
illustrated in \cref{fig:augmentation_methods}.
To create as diverse training samples as possible, we combine the augmentation methods by successively applying them to the same image. 
The key idea of our augmentation scheme is to randomly combine seven augmentation methods by applying each with a probability of $0.5$, after replacing the background. In that way, we not only apply multiple augmentation methods to the same image, but also vary the set of methods used. As a result, our augmentation scheme introduces much more variety than if certain methods were always applied in the exact same way and order.
This allows us to substantially improve the segmentation performance and the quality of the uncertainty quantification using deep ensembles when only a small non-representative dataset is available.

\begin{remark}
    While our proposed approach 
    can improve uncertainty quantification, it does not provide calibration guarantees. In order to obtain them, re-calibration techniques can be applied, e.g., by re-scaling the output of the deep ensemble such that obstacle pixels are guaranteed to be correctly classified with the desired probability or higher \cite{guo2017calibration}. Suitable scaling factors can be obtained from test images, e.g., using data-driven optimization approaches \cite{Campi2009}. 
    Alternatively, the empirical accuracy of the deep ensemble can be determined using test images, such that generalization guarantees such as \cite[Theorem 1]{Bradford2019} can be exploited to certify a reduced accuracy
    . Note that the results from both of these approaches are best when the calibration is already high, which underlines the importance of our proposed approach.\looseness=-1
\end{remark}


\section{Safe Vision-Based Motion Planning}
Based on the probabilistic semantic segmentation results, the goal is to plan and execute a $\delta$-safe motion as introduced in \cref{prob2}. 
To this end, we split the planning problem
into path planning and velocity scheduling. For solving the former, we present a scenario optimization approach in \cref{subsec:path_plan} and 
discuss its integration into the 
RRT* algorithm in \cref{subsec:rrt}.
Subsequently, we determine the maximum velocity profile along the obtained path still admitting a $\delta$-safe motion in \cref{subsec:trajectory_planning}.\looseness=-1

\subsection{Path Planning as Scenario Optimization Problem}\label{subsec:path_plan}

In order to admit a planning in the task space $\mathcal{T}$ of a robotic system, it is generally necessary to augment the probabilistic semantic segmentation result to 3D by using depth information, e.g., from a LiDAR scanner, a stereo camera system or available knowledge about the scene. {Potential errors in the depth measurements can be considered, e.g., by enlarging the obstacle accordingly in the direction of the camera.}
This allows us to compute the occupancy probability $P(\bm{x} \in \mathcal{W}_\mathrm{free})$ for each point $\bm{x}$ in the 3D workspace~$\mathcal{W}$. Moreover, we can determine the possibly occupied region ${\mathcal{R}(\bm{\pi}(t)) \oplus \mathcal{B}({\eta}(\gamma(\|\dot{\bm{\pi}}(t)\|){)})}$ in the workspace for each pose along a trajectory $\bm{\pi}(t)\in\mathcal{T}$, where $\mathcal{B}(r)$ is a sphere with radius $r$, $\gamma(\|\dot{\bm{\pi}}(t)\|)$ represents the tracking error bound defined in \cref{ass:track_error}{, and $\eta(\cdot)$ maps the task space error to the maximum resulting workspace error. The function $\eta$ can be computed based on the Jacobian mapping task space velocities to the work space. The computation of $\eta$ simplifies and the conservatism reduces if the tracking error in the task space orientation can be neglected.} 
The condition for a pose $\bm{\pi}(t)\in\mathcal{T}$ being $\delta$-safe can be expressed as 
\begin{align}
    P(\bm{x} \in \mathcal{W}_\mathrm{free}) \geq 1-\delta, \forall \bm{x} \in \mathcal{R}(\bm{\pi}(t)) \oplus \mathcal{B}({\tilde{\gamma}}(\|\dot{\bm{\pi}}(t)\|)){,}
\end{align}
{where $\tilde{\gamma}(\cdot)\defeq\eta(\gamma(\cdot))$.} As the dependence on the {reference} velocity $\dot{\bm{\pi}}(\cdot)$ is not suitable for standard path planning algorithms, we substitute it with a desired minimum velocity $\underline{v}\in\mathbb{R}_+$. 
This allows us to split the safe motion planning problem into a simple path planning problem followed by a velocity scheduling problem.

For {path} planning with uncertainty, strong assumptions are usually made about the shape of obstacles and their probability distribution to derive analytic expressions for the constraint in \cref{prob2} \cite{Zhu2019, Park2018, Kamel2017}. 
To avoid the associated conservatism {and directly use the uncertainty estimates obtained from the solution of \cref{prob1} for path planning}, we reformulate \cref{prob2} as a scenario problem \cite{Campi2009}. 
The path is discretized into $K\in\mathbb{N}$ poses $(\bm{p}_1, \dots, \bm{p}_K)$ and a fixed set $\mathcal{R}_0 \defeq \mathcal{R}(\bm{0})$ is defined, such that for all poses $\bm{p} \in \mathcal{T}$, the set $\mathcal{R}(\bm{p})$ can be described as a rigid motion $T^{\bm{p}}(\cdot)$ of $\mathcal{R}_0$, i.e., $\mathcal{R}(\bm{p}) = T^{\bm{p}}(\mathcal{R}_0)$.  This allows us to approximate
the safety constraint 
using random samples $\bm{x}^{(n,k)} = T^{\bm{p}_k}\big(\bm{x}_0^{(n)}\big)$, 
where $N_x\in\mathbb{N}$ vectors $\bm{x}_0^{(n)}$ are drawn from a uniform distribution $\mathcal{U}(\mathcal{R}_0 \oplus \mathcal{B}({\tilde{\gamma}}(\underline{v})))$.
This is a common approach for reformulating robust into scenario constraints \cite{Campi2009}.
It leads to the scenario optimization problem\looseness=-1
\begin{subequations}
\label{eq:path_plan_problem}
\begin{align}
    \min_{(\bm{p}_1, \dots, \bm{p}_K)} \:&\sum\nolimits_{j=1}^{K} c(\bm{p}_{{j}}) \\
    \label{eq:safety_constraint}
    \text{s.t.}\quad\; &P\big(\bm{x}^{(n,k)} \in \mathcal{W}_\mathrm{free}\big) \geq 1-\delta, \\ &\forall n = 1,\dots,N_x, \: k = 1,\dots,K, \nonumber
\end{align}
\end{subequations}
with $\bm{x}^{(n,k)} = T^{\bm{p}_k}\big(\bm{x}_0^{(n)}\big), \: \bm{x}_0^{(n)} \sim \mathcal{U}(\mathcal{R}_0 \oplus \mathcal{B}({\tilde{\gamma}}(\underline{v})))$. 
{While the planned path is directly affected by the choice of $\delta$ through~\eqref{eq:safety_constraint}, reducing $\delta$ can have different impact around the obstacle, as the extent of the uncertain area usually varies along the obstacle boundary.}
In problem~\eqref{eq:path_plan_problem}, the safety constraint is only evaluated at discrete points $\bm{p}_1,\dots,\bm{p}_k$ on the path. Still, $\delta$-safety of the continuous path can be achieved by enlarging the set ${\mathcal{R}_0  \oplus \mathcal{B}({\tilde{\gamma}}(\underline{v}))}$, accordingly. {For simple calculation, enlarging its enclosing sphere by $\mathcal{B}(l/2)$, where $l$ is the distance between the positions occupied at the ends of the line segment, can be performed. }
Based on the solution $(\bm{p}_1^*,\ldots,\bm{p}_K^*)$ of problem~\eqref{eq:path_plan_problem}, we define the continuous path $\bm{\pi}^*:[0,1] \mapsto \mathcal{T}$ by linearly interpolating between all $\bm{p}_k^*$ such that $\bm{\pi}^*(k\Delta s) = \bm{p}_{k+1}$ for $\Delta s=1/(K-1)$. 
Since the reliability of the scenario approximation grows with the number of random samples $N_x$ \cite{Campi2009}, this approach provides a well-suited obstacle representation for the proposed uncertainty-aware visual perception.\looseness=-1

\subsection{Path Planning with the SCC-RRT* Algorithm}\label{subsec:rrt}
For solving the scenario optimization problem~\eqref{eq:path_plan_problem}, we exemplarily employ a modified version of the popular RRT* path planning algorithm \cite{Karaman2011}, which we refer to as scenario chance-constrained RRT* (SCC-RRT*).
The collision checking represents the main difference between our SCC-RRT* algorithm and previous uncertainty-aware RRT* variants~\cite{Luders2013}. 
In order to evaluate whether a line segment $[\bm{p}_1, \bm{p}_2] \subset \mathcal{T}$ is eligible, the safety constraint \eqref{eq:safety_constraint} is evaluated for discrete poses ${\bm{p}^{(i)} = \bm{p}_1 + i \Delta_p \nicefrac{\bm{p}_2 - \bm{p}_1}{\|\bm{p}_2 - \bm{p}_1\|_2}}$, $ {i=0,1,\dots,\left\lceil\nicefrac{\|\bm{p}_2 - \bm{p}_1\|}{\Delta_p}\right\rceil}$, where $\Delta_p \in \R_+$. For each $\bm{p}^{(i)}$, we draw $N_x$ samples $\bm{x}_0^{(n)},\; n=1,\dots,N_x$, uniformly from the set $(\mathcal{R}_0 \oplus \mathcal{B}({\tilde{\gamma}}(\underline{v})))$ and apply the rigid motion $T^{\bm{p}^{(i)}}(\cdot)$ to each $\bm{x}_0^{(n)}$. If all points $T^{\bm{p}^{(i)}}(\bm{x}_0^{(n)})\in \mathcal{W}_\mathrm{free}$ lie inside the free workspace $\mathcal{W}_\mathrm{free}$ with probability at least $1-\delta$, the line segment $[\bm{p}_1,\bm{p}_2]$ is considered $\delta$-safe.
The quality of a path is evaluated using a line cost function. 
By employing the common RRT* rewiring procedure, our SCC-RRT* algorithm asymptotically converges to the minimum-cost path if the cost function is monotonous and bounded \cite{Karaman2011}.
{As our planning algorithm is global, the risk of getting stuck in local minima due to the unstructured nature of the occupancy estimates obtained from solving~\cref{prob1} is avoided.}
\looseness=-1

\subsection{Safe Velocity Scheduling}\label{subsec:trajectory_planning}

While the path obtained using \eqref{eq:path_plan_problem} admits a trajectory executed at the specified minimum velocity $\underline{v}$, a faster execution is possible when the obstacles are sufficiently far away. 
To find {an upper bound $v^*(s)$ on the velocity} for $\delta$-safety at some point $\bm{\pi}^*(s)$,
we need to determine the distance between the robot and the closest point which is not $\delta$-safe. This distance can be compactly expressed as  
\begin{subequations}
\label{eq:velocity_scheduling_distance}
\begin{align}
    d_{\mathrm{o}}(\bm{\pi}(s)) = \min\limits_{\bm{x}_\mathrm{r},\bm{x}_o} \: &\|\bm{x}_\mathrm{r}-\bm{x}_\mathrm{o}\|_2 \\
    \text{s.t.} \;\: &\bm{x}_\mathrm{r} \in \mathcal{R}(\bm{\pi}(s)) \\ &P(\bm{x}_\mathrm{o} \in \mathcal{W}_\mathrm{free}) < 1-\delta,
\end{align}
\end{subequations}
which can be effectively approximated by sampling multiple positions $\bm{x}_{\mathrm{o}}$ and merely optimizing with respect to $\bm{x}_\mathrm{r}$. 
Using the distance $d_{\mathrm{o}}(\bm{\pi}(s))$, we can easily determine $v^*(s)$ since ${\tilde{\gamma}}(\dot{\bm{\pi}}(s))\leq d_{\mathrm{o}}(\bm{\pi}(s))$ must hold for $\delta$-safety. Since the chain rule $\dot{\bm{\pi}}(s) = \frac{\diff\bm{\pi}(s)}{\diff s}\,\dot{s} = \bm{\pi}'(s)\dot{s}$ admits a parameterization of the velocity in terms of $\dot{s}$, we can obtain $v^*(s)$ by solving 
\begin{subequations}
\label{eq:ds_limit}
\begin{align}
    v^*(s) = \max_{a \geq 0} \; &\bm{\pi}'(s)a \\
    \text{s.t.} \;\: &{\tilde{\gamma}}(\|\bm{\pi}'(s)\|a) \leq d_{\mathrm{o}}(\bm{\pi}(s)) \\
    &\|\bm{\pi}'(s)\|a \leq \bar{v},
\end{align}
\end{subequations}
where $\bar{v}$ denotes the maximum executable velocity of the robot. 
{Unlike path planning, velocity scheduling is not directly affected by $\delta$, as \eqref{eq:velocity_scheduling_distance} is solved only for poses on the path already determined based on $\delta$.}
We discretize the range of $s\in[0,1]$ into $l$ steps 
and use a line-search to efficiently compute \eqref{eq:ds_limit} at the discretization points.
Finally, numerical integration of the obtained velocity profile yields a smooth $\delta$-safe trajectory~$\bm{\pi}^*(t)$ \cite{Bobrow1985}.

\section{Evaluation}
We evaluate the visual perception and motion planning modules separately and deploy the proposed framework for a real-world task. 
First, we examine the effect of data augmentation and 
ensembling 
in \cref{subsec:eval_aug}. 
As a segmentation model, we employ the DeepLabv3 architecture \cite{Chen17} with the ResNet50 CNN \cite{He16} as the backbone. 
In \cref{subsec:eval_plan}, we show the high flexibility of the proposed scenario formulation for planning in uncertain environments. Finally, in \cref{subsec:exp}, the effectiveness of the safe vision-based motion planning framework is demonstrated 
in an experiment with a 7 DOF robotic manipulator and a hand as obstacle.


\subsection{Uncertainty Quantification in Visual Perception}\label{subsec:eval_aug}

\begin{figure*}
    \centering
	\begin{subfigure}[b]{0.48\textwidth}
		\centering
		\includegraphics[width=\textwidth]{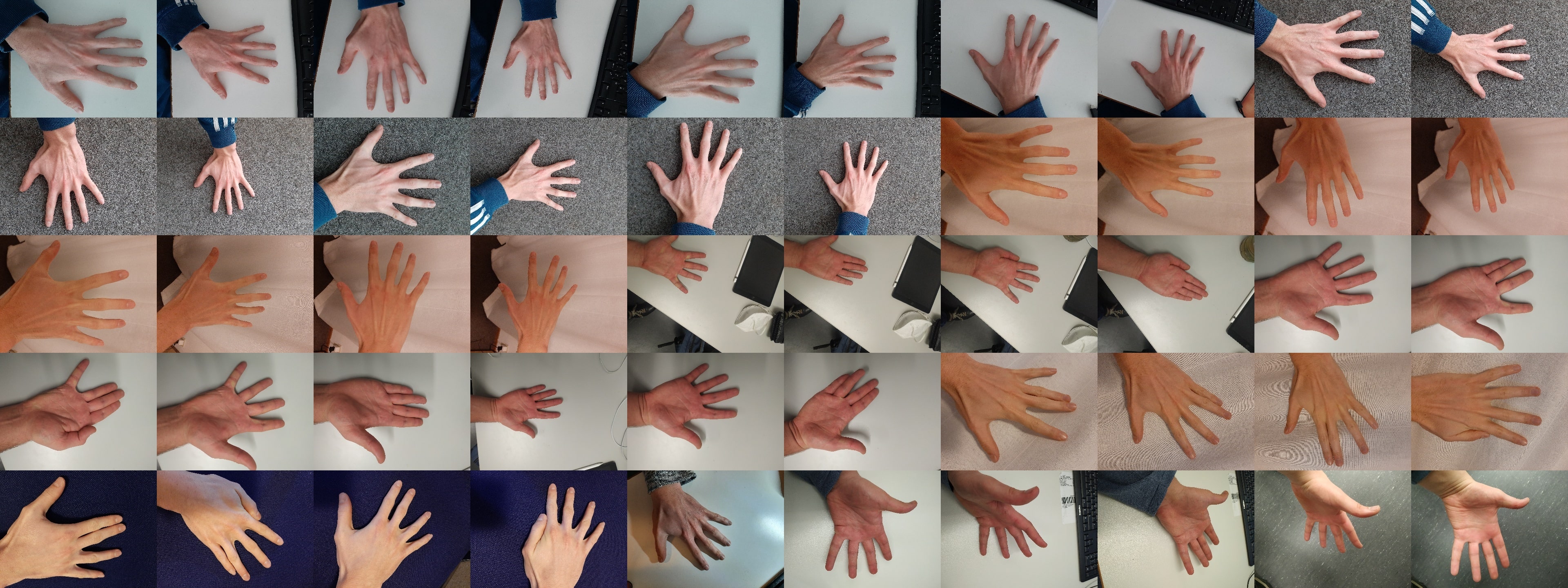}\\
		\caption{50 Training images}
		\label{subfig:trainingset}
	\end{subfigure}
	\hfill
	\begin{subfigure}[b]{0.48\textwidth}
		\centering
		\includegraphics[width=\textwidth]{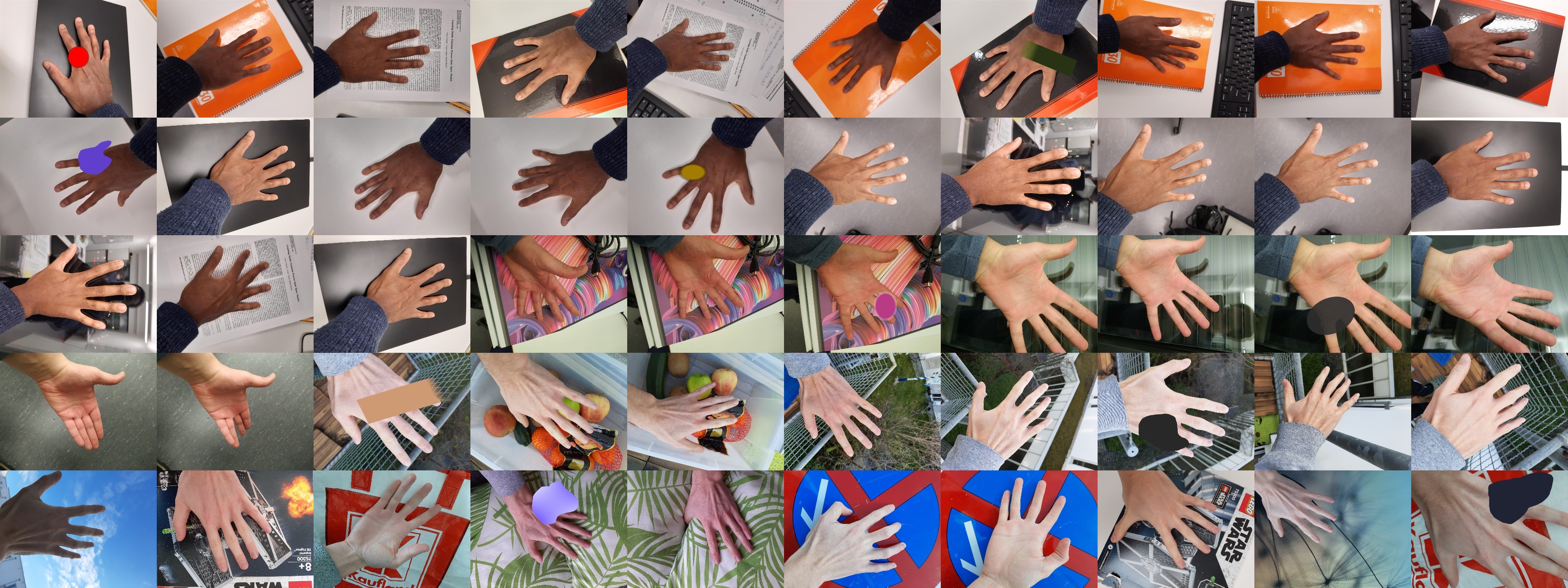}\\
		\caption{50 Test images}
		\label{subfig:testset}
	\end{subfigure}
	\caption{\small We split our available annotated samples into a training and a test set. The former contains only little variety and is thus not representative for the test set, making the training task a good example for \cref{prob1}.}
	\label{fig:dataset}
\end{figure*}

In accordance with \cref{prob1}, we aim to evaluate our uncertainty-aware visual perception approach for a small dataset with low variety containing images of complex-shaped objects. To this end, we consider highly accurate semantic segmentation of human hands and create our own dataset for this task\footnote{The images in the dataset show only hands of co-authors, who consented to the usage and publication.}.
Since the labelling procedure is time consuming, only 100 images are created and divided {equally} into a training and a testing set as depicted in Figures~\ref{subfig:trainingset} and \ref{subfig:testset}. 
It can clearly be seen that the training set is chosen such that it exhibits only limited diversity, e.g., all images have monotonous and very similar backgrounds. In contrast, the test set includes highly diverse images with hands from multiple people in various environments. Therefore, the training set is not representative of the test images, such that the resulting segmentation task is a good instance for \cref{prob1}, and thus well-suited to demonstrate the effectiveness of the proposed massive augmentation procedure in adding missing variability to a dataset.


\begin{table}[t]
\setlength{\tabcolsep}{3.7pt}
\centering \scriptsize
\vspace{0.2cm}
\begin{tabular}{ c c c c } 
    \hline
	Training set & Methods & PA $\pm$ $1$ std $[\%]$& mIoU  $\pm$ $1$ std $[\%]$ \\
    \hline
    \multirow{5}{*}{200 images} & Cutout \cite{Devries2017} & $88.2 \pm 0.9 $ & $78.7 \pm 1.4 $ \\ 
    & Mixup \cite{Zhang2018} & $89.4 \pm 1.4 $ & $78.3 \pm 2.0$ \\ 
    & Flip. + $90^\circ$ Rot. \cite{Nanni2021} & $88.5 \pm 1.0$ & $79.4 \pm 1.5$ \\ 
    & Flip. + Rot. + Crop. \cite{Uzun2021} & $86.9 \pm 1.3 $ & $76.7 \pm 1.9 $ \\ 
    & Our scheme & $\bm{95.4 \pm 0.7}$ & $\bm{89.1 \pm 1.4}$ \\ 
    \hline
    \multirow{5}{*}{1000 images} & Cutout \cite{Devries2017} & $91.7 \pm 0.6 $ & $83.6 \pm 0.8$ \\ 
    & Mixup \cite{Zhang2018} & $88.1 \pm 0.8 $ & $77.1 \pm 1.1 $ \\ 
    & Flip. + $90^\circ$ Rot. \cite{Nanni2021} & $88.2 \pm 1.0 $ & $78.1 \pm 1.4 $ \\ 
    & Flip. + Rot. + Crop. \cite{Uzun2021} & $89.2 \pm 1.1$ & $79.7 \pm 1.7$ \\ 
    & Our scheme & $\bm{96.2 \pm 0.8}$ & $\bm{91.0 \pm 1.6}$ \\ 
    \hline
    \end{tabular}\vspace{-0.2cm}
	\captionof{table}{\small Our augmentation scheme significantly outperforms other methods for creating datasets with small (200) and large (1000) size. 
	}
	\label{tab:aug_methods_comb}
\end{table}


\begin{table}
\setlength{\tabcolsep}{3.7pt}
\centering \scriptsize
\begin{tabular}{ c c c c } 
    \hline
	Training set & Methods & NLL $\times 10^2$ & BS $\times 10^3$ \\
    \hline
    \multirow{5}{*}{200 images} & Cutout \cite{Devries2017} & $88.1$ & $41.8$ \\ 
    & Mixup \cite{Zhang2018} & $68.3$ & $25.0$ \\ 
    & Flip. + $90^\circ$ Rot. \cite{Nanni2021} & $91.7$ & $47.5$ \\ 
    & Flip. + Rot. + Crop. \cite{Uzun2021} & $97.4$ & $41.2$ \\ 
    & Our scheme & $\bm{25.4}$ & $\bm{10.5}$ \\ 
    \hline
    \multirow{5}{*}{1000 images} & Cutout \cite{Devries2017} & $63.1$ & $32.3$ \\ 
    & Mixup \cite{Zhang2018} & $87.1$ & $36.9$ \\ 
    & Flip. + $90^\circ$ Rot. \cite{Nanni2021} & $84.1$ & $47.0$ \\ 
    & Flip. + Rot. + Crop. \cite{Uzun2021} & $80.2$ & $39.4$ \\ 
    & Our scheme & $\bm{23.4}$ & $\bm{11.4}$ \\ 
    \hline
    \end{tabular}\vspace{-0.2cm}
    \captionof{table}{\small Data augmentation has a significant impact on uncertainty quantification. The impact of the size of the training set depends on the used approach for data augmentation.
    }
    \label{tab:bs_nll_app}
\end{table}

We evaluate the combination of the data augmentation methods within our augmentation scheme. For this purpose, we apply our scheme and the methods proposed in \cite{Devries2017, Zhang2018, Uzun2021, Nanni2021} to create two augmented datasets 
{from the 50 training images}: A small one containing 200 images and a large one containing 1000 images. 
In \cref{tab:aug_methods_comb}, the results for our scheme in terms of pixel accuracy (PA) and mean intersection-over-union (mIoU) \cite{Shelhamer17} are presented together with the results of the comparison methods. It can be seen that our scheme provides significantly stronger improvements in model performance for both degrees of dataset inflation. The best performance is achieved when {the dataset is enlarged from 50 to 1000 images using our scheme, i.e., by a factor of 20}, which is ten times {the factor of two used} in \cite{Uzun2021}. 
We also examine the impact of our data augmentation scheme on calibration and compare it to the approaches from \cite{Devries2017, Zhang2018, Uzun2021, Nanni2021} for an inflated training set of 200 and 1000 images, respectively. To this end, we consider two common metrics, the Brier score (BS) and negative log-likelihood (NLL) \cite{Lakshminarayanan17}.
As shown in Table~\ref{tab:bs_nll_app}, our augmentation scheme yields much better predictive uncertainty estimates than the comparison methods, including mixup, which is a popular method to improve calibration \cite{Zhang2018}
. Our results show the importance of addressing the lacking variety within the training set through massive data augmentation combining different augmentation methods. 

\begin{table}[t]
    \scriptsize
    \setlength{\tabcolsep}{3pt}
    \centering
    \begin{tabular}{ c c c c c c c c c c c} 
		\hline
		M & 1 & 2 & 3 & 4 & 5 & 6 & 7 & 8 & 9 & 10\\
		\hline
		BS $\times 10^3$ & 30.8 & 29.7 & 28.1 & 26.9 & 25.1 & 24.3 & 23.9 & 23.3 & 23.6 & 23.4 \\
		NLL $\times 10^2$ & 18.0 & 16.1 & 13.9 & 13.0 & 12.3 & 12.1 & 11.9 & 11.7 & 11.4 & 11.4 \\
		\hline
	\end{tabular}
	\captionof{table}{\small Predictive uncertainty improves with increasing ensemble size $M$, which is shown by decreasing Brier score (BS) and negative log-likelihood (NLL).}
	\label{tab:bs_nll}
\end{table}

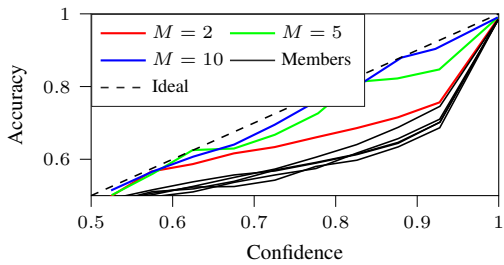
\begin{figure}
  \centering
\begin{tikzpicture}
\begin{axis}[
	width=7cm,
	height=4.cm,
	tick align=outside,
	tick pos=left,
	x grid style={white!69.0196078431373!black},
	xmin=0.5, xmax=1,
	xlabel={\footnotesize Confidence}, xlabel style={align=center},
	ylabel style={align=center, yshift=-0.25cm}, ylabel={\footnotesize Accuracy},
	xtick style={color=black},
	y grid style={white!69.0196078431373!black},
	ymin=0.5, ymax=1,
	ytick style={color=black},
	legend columns=2,
	legend style={at={(0,1)}, anchor=north west, legend cell align=left, align=left, draw=white!15!black, font=\scriptsize},
	x tick label style={font=\scriptsize}, y tick label style={font=\scriptsize}
	]
	\addplot [thick, red]
	table {%
0.513291226964217 0.48305519736595
0.574028295693021 0.566366258111031
0.624550231856343 0.586642824468749
0.675061306893681 0.616182572614108
0.725193277746545 0.633545720681936
0.775508351379152 0.660092122357387
0.825720723497382 0.685692157456982
0.876242934334126 0.715319536639334
0.927413766088543 0.756456229233148
0.999170389932899 0.986689980086454
	};
	\addlegendentry{$M=2$}
	
	\addplot [thick, green]
	table {%
0.525113008955515 0.501163692785105
0.575919316300724 0.560701589643839
0.624548256584517 0.62524153033621
0.675159228623249 0.629497345828691
0.725425555216411 0.667466425700308
0.778383964334037 0.726394180065001
0.823708197398596 0.812396694214876
0.875647108559474 0.822292727145945
0.927169604372445 0.847084445339644
0.998924001207514 0.990130155509489
	};
	\addlegendentry{$M=5$}
	
	\addplot [thick, blue]
	table {%
0.524459480957072 0.514287565606169
0.57531337901951 0.566146450678256
0.625343145157696 0.607270522149879
0.675366832838566 0.640405226838937
0.725179233579478 0.695007255334014
0.777114836874197 0.760723296888141
0.824410718458635 0.798152382694594
0.880732565801582 0.879634868134738
0.922348985795828 0.903560913282595
0.998860614949279 0.990218575444395
	};
	\addlegendentry{$M=10$}
	
	\addplot [semithick]
	table {%
0.525108049449517 0.483213508343777
0.575249073964184 0.510896979887473
0.62540592070373 0.518537200504414
0.675362798118447 0.539335196527324
0.7254540281346 0.570210659479677
0.775435690372662 0.59114632757639
0.82592359841859 0.614334606813117
0.876420359044259 0.646299433378118
0.927988055030917 0.70182049693716
0.999190452782784 0.984582225853412
	};
	\addlegendentry{Members}
	
	\addplot[semithick, dashed]
	table{
	0.5 0.5
	1 1
	};
	\addlegendentry{Ideal}
	
	\addplot [semithick]
	table {%
0.525041435904056 0.483703656272011
0.57501223897626 0.506742281033396
0.625068620119016 0.523656630520385
0.675428107338635 0.525015906067442
0.725458059892446 0.542533180060612
0.775514250620372 0.580215091991823
0.825992282862235 0.597402124872653
0.876654715448127 0.634030547190798
0.927647567305879 0.686665883261428
0.999127590292533 0.982983616454792
	};
	
	\addplot [semithick]
	table {%
0.524961331657578 0.487237977805179
0.574893217825127 0.496412501520126
0.625034288150338 0.510261522223525
0.675305899171082 0.53647854529327
0.725420455746546 0.556763639925556
0.775775434231822 0.574520770183153
0.825997794468409 0.617158103578026
0.876691605269934 0.656750749508942
0.927533723269849 0.710158890608712
0.999135328104417 0.982245004227835
	};
	
\addplot [semithick]
	table {%
0.52514598358516 0.484987063000181
0.575100187722688 0.502406494636126
0.625261451358169 0.525510204081633
0.6751096523054 0.549902253318243
0.725482647484975 0.57303534202744
0.775802794959655 0.605686290930408
0.825834397962285 0.640382440629176
0.876688950508363 0.6878738170347
0.927844342392972 0.745604680215729
0.999039232410127 0.984434129617755
	};
	
\addplot [semithick] 
	table {%
0.524953975145419 0.491281273692191
0.575428989980786 0.516021042563367
0.625210358197364 0.538029893383666
0.675307585308288 0.556854610597775
0.725573399861808 0.567902829822975
0.775779049472457 0.589427344830095
0.826035781974401 0.612876647834275
0.87653078776035 0.64384404173531
0.927645138505778 0.701870392342456
0.99891085108991 0.982735340878972
	};
\end{axis}
\end{tikzpicture}
    \captionof{figure}{\small The reliability curves illustrate the good calibration of ensembles trained with massive data augmentation, justifying the interpretation of the semantic segmentation output as probabilities.}
    \label{fig:reliability_curve}
\end{figure}

Our safe planning framework builds on the assumption that for each pixel, the segmentation output can be used as an occupa{ncy} probability. To justify this assumption, we first examine calibration via the reliability diagram
. For this, the confidence interval $[0.5, 1)$ is partitioned into ten equally sized bins. 
For each bin, the {pixel accuracy} is calculated and plotted against the average confidence value of the pixels within the bin. 
We evaluate three differently sized ensembles, $M\in\{2,5,10\}$, and the individual members of the medium-sized ensemble, which are trained with our augmented training dataset. The results depicted in Figure \ref{fig:reliability_curve} show that calibration improves with increasing ensemble size. Moreover, as shown in \cref{tab:bs_nll}, BS and NLL significantly decrease with increasing ensemble size, which is consistent with the results reported in \cite{Lakshminarayanan17}. 
The reliability curve as well as BS and NLL indicate only small improvements for $M > 5$. Therefore, we conclude an ensemble size of $M = 5$ to maintain a good balance between computational demand and calibration suitable for the considered visual perception task. 

\subsection{Uncertainty Representations for Collision Avoidance}\label{subsec:eval_plan}

We compare the performance of path planning with our perception-based collision avoidance with three popular methods \cite{Zhu2019, Park2018, Kamel2017} that are based on the common assumption of uncertain position $\bm{x}_\mathrm{o} \sim \mathcal{N}(\hat{\bm{x}}_\mathrm{o}, \sigma^2\bm{I}_3)$, and known geometry and orientation of the obstacles. We consider a 4D task space $\mathcal{T}$ composed of the robot position $\bm{x}\in \mathcal{W}\subset \R^3$ and its orientation $\phi$ around the vertical $z$-axis. The set $\mathcal{R}(\bm{p})$ is described as an ellipse parallel to the $x$-$y$-plane, we set $\underline{v}=0.01\, \nicefrac{\si{m}}{\si{s}}$, $\bar{v}=0.2\, \nicefrac{\si{m}}{\si{s}}$, ${{\tilde{\gamma}}(v)=v/\bar{v}\cdot 0.01\, \si{m}}$, and consider spheres and cuboids as obstacles.  
For applying our approach, we consider that the occupancy probability decreases linearly with the distance $d$ to the obstacle surface\footnote{We could just as well assume other profiles for the probabilistic segmentation around the object boundaries. However, many predictions we obtain when perceiving real objects show a roughly linear decrease.}, becoming zero at $d=d_\mathrm{stop}$. Since we employ the chance constraint with probability $\delta=0.05$, this leads to $\tilde{d}=0.95d_{\mathrm{stop}}$ representing the extended boundaries of the uncertain object. Due to the Gaussian distribution of the object positions, these boundaries are equivalently parameterized using $2\sigma$ in the existing approaches \cite{Zhu2019, Park2018, Kamel2017}. For comparing the different methods, we create a simple scene containing three large obstacles and a more cluttered scenario with eight small obstacles. 
For solving \eqref{eq:path_plan_problem}, we employ the SSC-RRT* algorithm proposed in \cref{subsec:rrt}. 
The number of search tree iterations is set to $N_{\mathrm{iter}}=2000$.
We aim for a path that is short in Cartesian space and also avoids unnecessary rotations. To this end, we define the line cost function for a line segment $[\bm{p}, \bm{p}']\subset \mathcal{T}$ between two poses $\bm{p} = \big[\bm{x}\tp, \phi\big]\tp$, $\bm{p}' = \big[{\bm{x}'}\tp, \phi'\big]\tp$ in the task space as
\begin{align}
    \label{eq:eval_cost_fct}
    c_\mathrm{l}(\bm{p}, \bm{p}') = \|\bm{x}-\bm{x}'\|^2 + r(\phi - \phi')^2,
\end{align}
where $r>0$ {determines} the penalty for orientation changes.

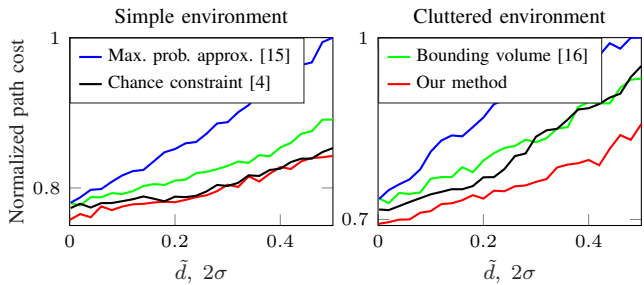
\begin{figure}
\centering
\setlength{\fwidth}{3.5cm}
\setlength{\fheight}{2.5cm}

\begin{tikzpicture}
\begin{axis}[%
width=\fwidth,
height=\fheight,
scale only axis,
xmin=0,
xmax=0.5,
xlabel style={font=\color{white!15!black},yshift = 0.2cm},
xlabel={\footnotesize $\tilde{d},\; 2\sigma$},
title={\footnotesize Simple environment},
title style={yshift=-0.2cm},
ymin=0.75,
ymax=1,
ytick={0.8, 1},
ylabel style={font=\color{white!15!black}, yshift=-0.55cm},
ylabel={\footnotesize Normalized path cost},
axis background/.style={fill=white},
legend style={at={(0,1)}, anchor=north west, legend cell align=left, align=left, draw=white!15!black, font=\scriptsize},
legend image post style={scale=0.5},
x tick label style={font=\scriptsize}, y tick label style={font=\scriptsize}
]
\addplot [color=red, thick, forget plot]
  table[row sep=crcr]{%
0	0.757430109478091\\
0.02	0.765269907993196\\
0.04	0.760978490061341\\
0.06	0.775420969331198\\
0.08	0.770498863716202\\
0.1	0.775259042680077\\
0.12	0.778215153334538\\
0.14	0.778780651115585\\
0.16	0.780665792341083\\
0.18	0.781635202688842\\
0.2	0.780995844350063\\
0.22	0.784079851423569\\
0.24	0.787711728855802\\
0.26	0.790258309534232\\
0.28	0.795703257025711\\
0.3	0.804282821263635\\
0.32	0.801275119267781\\
0.34	0.815416701379095\\
0.36	0.808538548059194\\
0.38	0.818988725405166\\
0.4	0.827764647892226\\
0.42	0.825207914880329\\
0.44	0.835375098541636\\
0.46	0.839990297222416\\
0.48	0.840747144178337\\
0.5	0.842573576006765\\
};

\addplot [color=green, thick, forget plot]
  table[row sep=crcr]{%
0.002	0.780028867526521\\
0.02	0.776708420724824\\
0.04	0.787702972623526\\
0.06	0.787649011954682\\
0.08	0.793027095592622\\
0.1	0.791971714224458\\
0.12	0.795733894176409\\
0.14	0.802489104881173\\
0.16	0.805117591781061\\
0.18	0.803986232833966\\
0.2	0.809773774510654\\
0.22	0.810980133601923\\
0.24	0.819661760280036\\
0.26	0.821365786767723\\
0.28	0.82502622617589\\
0.3	0.829239213038286\\
0.32	0.835040716529698\\
0.34	0.833286762616784\\
0.36	0.843660971019475\\
0.38	0.842116012538786\\
0.4	0.85373370992229\\
0.42	0.859886784289483\\
0.44	0.872067071793617\\
0.46	0.87548654892072\\
0.48	0.890777350735166\\
0.5	0.890854951626262\\
};

\addplot [color=blue, thick]
  table[row sep=crcr]{%
0.002	0.780323998618919\\
0.02	0.787384412718848\\
0.04	0.797469994909942\\
0.06	0.798516011829145\\
0.08	0.808192646150292\\
0.1	0.816664018779226\\
0.12	0.822349006042793\\
0.14	0.823932526212732\\
0.16	0.835565020882158\\
0.18	0.847433446041928\\
0.2	0.851749885712803\\
0.22	0.859326263048285\\
0.24	0.861081772549323\\
0.26	0.872206574223478\\
0.28	0.885757934761984\\
0.3	0.887474042371497\\
0.32	0.900723358923262\\
0.34	0.910299265029024\\
0.36	0.925458059311511\\
0.38	0.934360810564309\\
0.4	0.949568671523027\\
0.42	0.953407989637555\\
0.44	0.963567708667129\\
0.46	0.967015956805577\\
0.48	0.993895187830128\\
0.5	1\\
};
\addlegendentry{Max. prob. approx. \cite{Park2018}}

\addplot [color=black, thick]
  table[row sep=crcr]{%
0.002	0.773266420065988\\
0.02	0.778926802758577\\
0.04	0.773900190694465\\
0.06	0.779797748432034\\
0.08	0.779916982922879\\
0.1	0.782363024597763\\
0.12	0.784949251215788\\
0.14	0.788637784374276\\
0.16	0.785618296055705\\
0.18	0.782182483021081\\
0.2	0.788375440897693\\
0.22	0.787747942898609\\
0.24	0.789503977187306\\
0.26	0.795368358761674\\
0.28	0.803968659343247\\
0.3	0.802617039443205\\
0.32	0.806764269364824\\
0.34	0.816738094768669\\
0.36	0.815950514600084\\
0.38	0.823882781331804\\
0.4	0.825577589683492\\
0.42	0.834363876677654\\
0.44	0.838904509623018\\
0.46	0.839010631453768\\
0.48	0.847702153056576\\
0.5	0.853038671604628\\
};
\addlegendentry{Chance constraint \cite{Zhu2019}}
\end{axis}

\begin{axis}[%
xshift=4.1cm,
width=\fwidth,
height=\fheight,
scale only axis,
xmin=0,
xmax=0.5,
title={\footnotesize Cluttered environment},
title style={yshift=-0.15cm},
xlabel style={font=\color{white!15!black},yshift = 0.2cm},
xlabel={\footnotesize $\tilde{d},\; 2\sigma$},
ymin=0.69,
ymax=1,
ylabel style={font=\color{white!15!black}},
ytick={0.7, 1},
axis background/.style={fill=white},
legend style={at={(0,1)}, anchor=north west, legend cell align=left, align=left, draw=white!15!black, font=\scriptsize},
legend image post style={scale=0.5},
x tick label style={font=\scriptsize}, y tick label style={font=\scriptsize}
]
\addplot [color=green, thick]
  table[row sep=crcr]{%
0.002	0.734904164079835\\
0.02	0.726822973511107\\
0.04	0.743585034977358\\
0.06	0.742187914484697\\
0.08	0.744464751097605\\
0.1	0.767142788469443\\
0.12	0.769912497756434\\
0.14	0.769814641016846\\
0.16	0.786287391425688\\
0.18	0.777857497325341\\
0.2	0.797085516881564\\
0.22	0.809301081299516\\
0.24	0.81791660588676\\
0.26	0.821284193275706\\
0.28	0.831579627296735\\
0.3	0.827356562489756\\
0.32	0.834069647179468\\
0.34	0.849766169256842\\
0.36	0.852295634957384\\
0.38	0.885288561999544\\
0.4	0.893614647188946\\
0.42	0.891565210468201\\
0.44	0.891484390905976\\
0.46	0.916999077784292\\
0.48	0.930496219690845\\
0.5	0.932607860448141\\
};
\addlegendentry{Bounding volume \cite{Kamel2017}}

\addplot [color=red, thick]
  table[row sep=crcr]{%
0	0.692584725031886\\
0.02	0.695501389202223\\
0.04	0.699631000011804\\
0.06	0.70004142610954\\
0.08	0.711098945290775\\
0.1	0.713781062362603\\
0.12	0.725259388593202\\
0.14	0.726699654798248\\
0.16	0.731725789381374\\
0.18	0.740098936537168\\
0.2	0.734331359915491\\
0.22	0.747441417949241\\
0.24	0.746530526568912\\
0.26	0.754638742530897\\
0.28	0.756111256732704\\
0.3	0.761983170937351\\
0.32	0.766615994657492\\
0.34	0.785722432815451\\
0.36	0.79045128848555\\
0.38	0.792725337590485\\
0.4	0.79810533910454\\
0.42	0.789435048062245\\
0.44	0.816490627699057\\
0.46	0.839221642444171\\
0.48	0.831301448754663\\
0.5	0.857401614445154\\
};
\addlegendentry{Our method}

\addplot [color=blue, thick]
  table[row sep=crcr]{%
0.002	0.733768131497498\\
0.02	0.748315597528719\\
0.04	0.757851665603519\\
0.06	0.766216661003003\\
0.08	0.782623247067044\\
0.1	0.811781031331923\\
0.12	0.830098981832356\\
0.14	0.838632918387418\\
0.16	0.837151517442624\\
0.18	0.852365796688676\\
0.2	0.868101267850566\\
0.22	0.890111129582003\\
0.24	0.896826137111895\\
0.26	0.909026768227949\\
0.28	0.917872897575315\\
0.3	0.938544319201793\\
0.32	0.939957548710858\\
0.34	0.947965336883294\\
0.36	0.954994226660214\\
0.38	0.95352988408485\\
0.4	0.961180698656454\\
0.42	0.970353777420156\\
0.44	0.990774261146127\\
0.46	0.981755941294934\\
0.48	0.999659547891646\\
0.5	1\\
};

\addplot [color=black, thick]
  table[row sep=crcr]{%
0.002	0.71642724767175\\
0.02	0.715372829074321\\
0.04	0.722003177238574\\
0.06	0.728824274860405\\
0.08	0.735071410474004\\
0.1	0.74140591111206\\
0.12	0.745455308467086\\
0.14	0.749910873577116\\
0.16	0.749826266863926\\
0.18	0.755080802489457\\
0.2	0.76949711553589\\
0.22	0.769809845502495\\
0.24	0.78313233323873\\
0.26	0.80455016074461\\
0.28	0.808780643462338\\
0.3	0.836561913634849\\
0.32	0.847008567838723\\
0.34	0.849497480035221\\
0.36	0.864754296165787\\
0.38	0.881363169422107\\
0.4	0.883522234028099\\
0.42	0.890413908330681\\
0.44	0.901011707206929\\
0.46	0.907485883467204\\
0.48	0.934865556190277\\
0.5	0.953050990091342\\
};
\end{axis}
\end{tikzpicture}
\captionof{figure}{We compare different uncertainty-aware collision avoidance strategies for path planning within the CC-RRT* framework. Our perception-based approach leads to less conservative paths, with the advantage becoming stronger for more cluttered environments.}
\label{fig:planning_comp}

\end{figure}
We run the SCC-RRT* algorithm using this cost function and $N_x=100$ for $100$ times per method and scene over a range of different values of $\tilde{d}$ and $2\sigma$, which represent different uncertainty levels in the perception. The resulting normalized average costs are illustrated in \cref{fig:planning_comp}. 
While the behavior of the cost for the simple scenario is almost identical for our method and the approach proposed in \cite{Zhu2019}, the cost achieved using the existing methods \cite{Zhu2019, Park2018, Kamel2017} increases significantly faster in the cluttered environment. The proposed SCC-RRT* algorithm achieves this slower deterioration of the planning performance through the sampling-based obstacle representation, which ensures that our approach does not directly depend on the number of obstacles. This is in contrast to parametric obstacle representations in existing methods \cite{Zhu2019, Park2018, Kamel2017}, where the conservatism of the uncertainty approximations for the individual obstacles accumulates, such that the path performance crucially suffers from growing uncertainties as measured by~$\sigma$.\looseness=-1

\begin{figure}
  \centering
  \vspace{0.2cm}
\begin{tikzpicture}
\begin{axis}[
	width=8.5cm,
	height=4.2cm,
	tick align=outside,
	tick pos=left,
        ymode=log,
        log ticks with fixed point,
	x grid style={white!69.0196078431373!black},
	xmin=1, xmax=100,
	xlabel={\footnotesize Number of obstacles}, xlabel style={align=center},
	ylabel style={align=center, yshift=-0.25cm}, ylabel={\footnotesize Computation time [\si{s}]},
        xtick={1, 20, 40, 60, 80, 100},
	xtick style={color=black},
	y grid style={white!69.0196078431373!black},
	ymin=2, ymax=350,
	ytick style={color=black},
        extra y ticks={2},
        extra y tick labels={2},
	legend columns=2,
	legend style={at={(1,1)}, anchor=north east, legend cell align=left, align=left, draw=white!15!black, font=\scriptsize},
        legend image post style={scale=0.5},
	x tick label style={font=\scriptsize}, y tick label style={font=\scriptsize}
	]
        \addplot [thick, blue]
	table[row sep=crcr] {%
            1	30.15820148\\
            2	70.13762391\\
            3	113.51435796\\
            4	146.14379706\\
            5	177.20340751\\
            6	225.69251996\\
            7	250.1374673\\
            8	302.16131817\\
            9	309.96552222\\
            10	348.66515125\\
	};
	\addlegendentry{Max. prob. approx. \cite{Park2018}} 

        \addplot [thick, black]
	table[row sep=crcr] {%
            1	3.71358017\\
            2	4.28577875\\
            3	4.84138763\\
            4	5.387057\\
            5	5.87540757\\
            6	6.48462384\\
            7	7.10069043\\
            8	7.78509744\\
            9	8.29913161999999\\
            10	8.97004004\\
            15	13.10971059\\
            20	16.63242109\\
            25	18.21205125\\
            30	22.70944839\\
            35	23.87488404\\
            40	25.9805041\\
            45	30.62263381\\
            50	35.02444943\\
            55	34.23270433\\
            60	37.2205743\\
            65	42.08253369\\
            70	41.77974121\\
            75	46.25718826\\
            80	48.09613844\\
            85	51.13381321\\
            90	51.71578684\\
            95	57.86831856\\
            100	55.96687877\\
	};
	\addlegendentry{Chance constraint      \cite{Zhu2019}} 

        \addplot [thick, green]
        table[row sep=crcr]{%
            1	3.06954169\\
            2	3.13872467\\
            3	3.30103597\\
            4	3.52585843\\
            5	3.55240078\\
            6	3.64031419\\
            7	4.30923406\\
            8	3.91280254\\
            9	5.8773219\\
            10	5.7382104\\
            15	5.48035398\\
            20	7.36213422\\
            25	6.73410392\\
            30	7.84558954\\
            35	8.41739706\\
            40	9.01956334\\
            45	10.42185543\\
            50	10.23422303\\
            55	10.15081896\\
            60	10.66127012\\
            65	11.255745\\
            70	12.17663578\\
            75	12.58006657\\
            80	11.90282792\\
            85	12.51537698\\
            90	14.80499265\\
            95	14.52478325\\
            100	15.23865224\\
        };
        \addlegendentry{Bounding volume \cite{Kamel2017}}

        \addplot [thick, red]
        table[row sep=crcr]{%
            1	8.23477643\\
            100	8.23477643\\
        };
        \addlegendentry{Our method} 
        
        \addplot [thick, red]
        table[row sep=crcr]{%
            1	4.56188 \\
            100	4.56188 \\
        };
        
        \node[] at (axis cs: 90, 6.5) {\scriptsize $N_x = 100$};
        \node[] at (axis cs: 89, 3.5) {\scriptsize $N_x = 20$};

\end{axis}
\end{tikzpicture}
    \captionof{figure}{\small The average computation times for the RRT* algorithm with parametric collision checking increase significantly with the number of obstacles. In contrast, for our vision-based obstacle representation, the computation time only depends on the number of samples $N_x$, which is especially beneficial in cluttered environments.}
    \label{fig:computation_times}
\end{figure}
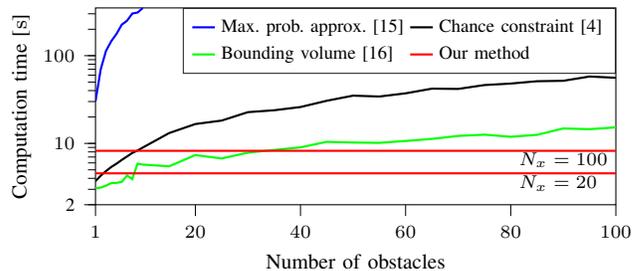

Additionally, we compare the computational efficiency of the proposed scenario approach with existing parametric approaches. For this, we create scenes with different numbers of randomly placed spherical obstacles. The average computation times for running the RRT* algorithm 100 times with the collision checking methods \cite{Zhu2019, Park2018, Kamel2017} are recorded and shown in \cref{fig:computation_times}. For comparison, we also include the average computation time of our sampling-based approach for different values of $N_x$. The parametric methods exhibit a significant increase in computation time with the number of obstacles, which must all be checked for collision.
The SCC-RRT* algorithm does not suffer from this issue as it exploits a joint environment representation that is only sampled at test points instead of individual object representations.
Thus, the proposed scenario approach simultaneously achieves flexibility and efficiency and is applicable to highly cluttered environments containing many obstacles.

\subsection{Experimental Evaluation with a KUKA iiwa robot}\label{subsec:exp}
\begin{figure}
  \centering
  \includegraphics[width=0.4\linewidth]{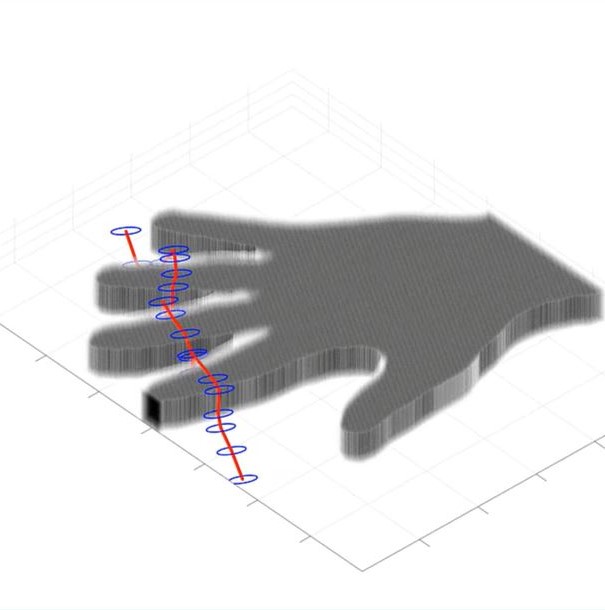}
  \hspace{0.9cm}
  \includegraphics[width=0.4\linewidth]{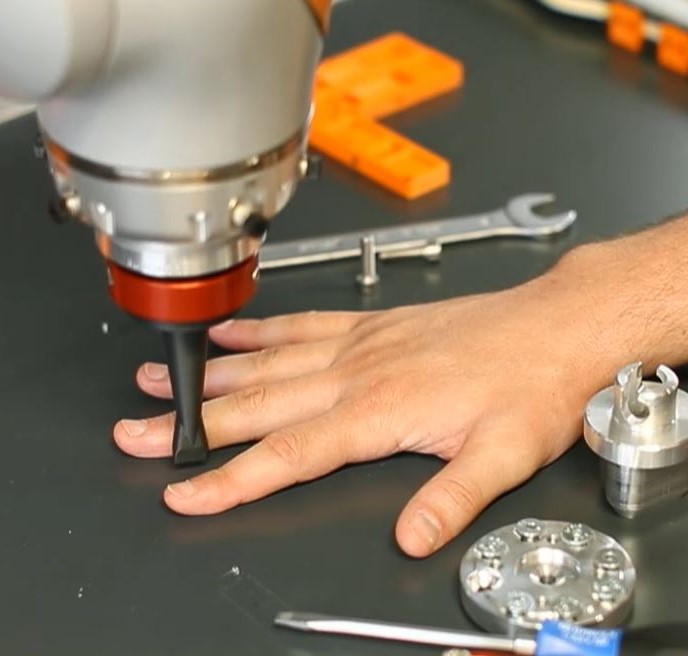}
  \captionof{figure}{We experimentally validate our approach with a robotic manipulator.}
  \label{fig:experiment}
\end{figure}
\begin{figure}[t]
  \centering
  \includegraphics[width=\linewidth]{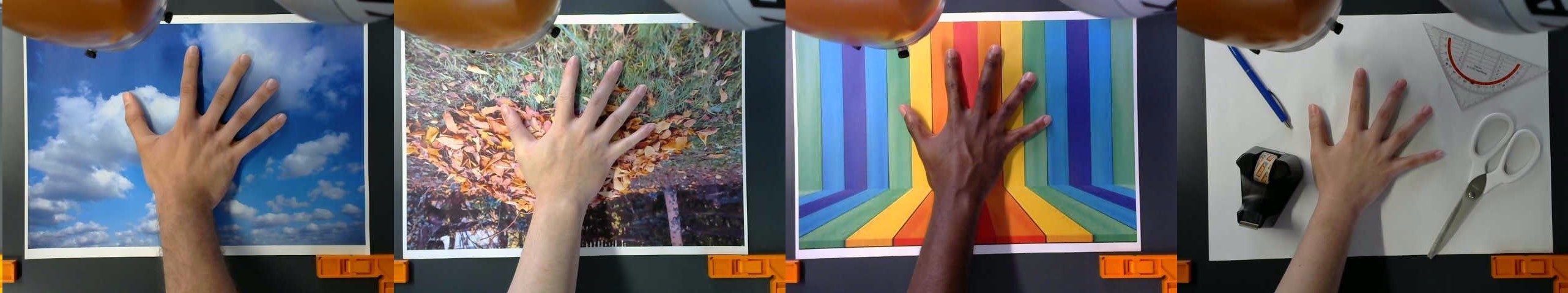}
  \captionof{figure}{The experiment is conducted with different participants, lighting conditions and backgrounds, which are not represented by our training data.}
  \label{fig:experiment_scenes}
\end{figure}
In order to show the real-world applicability 
of our approach, we conduct an experiment with an impedance controlled KUKA iiwa robotic manipulator with seven degrees of freedom and a hand as obstacle\footnote{Approval for this type of experiments involving the close interaction between a Kuka iiwa manipulator and a human has been obtained by the  ethics committee of the medical faculty of the Technical University of  Munich.\looseness=-1}. 
A Logitech C270 USB webcam positioned 0.5 m above the working area takes images of the workspace. We apply the deep ensemble trained as discussed in \cref{subsec:eval_aug} and augment the 2D probabilistic semantic segmentation result to 3D by assuming fixed hand height. We use the same expression for $\underline{v}$, $\bar{v}$ and $\mathcal{R}(\bm{p})$ as in Section~\ref{subsec:eval_plan} and plan a $\delta$-safe path with the SCC-RRT* algorithm with $N_\mathrm{iter}=2000$. {From tracking experiments with the manipulator, we obtain an approximate tracking error bound ${{\tilde{\gamma}}(v)=v/\bar{v}\cdot 0.01\, \si{m}}$. This procedure requires test trajectories that cover a sufficiently wide range of operating conditions, which can be challenging for higher-dimensional task spaces.}
To obtain a motion in close proximity to the fingers, we replace the first term in the cost function~\eqref{eq:eval_cost_fct} with a term that penalizes the area between the path and the table. The motion is executed with a maximum velocity $\bar{v}=0.2\, \nicefrac{\text{m}}{\text{s}}$, see Figure~\ref{fig:experiment}. As shown in Figure~\ref{fig:experiment_scenes} and in our supplementary video
, the experiment is successfully repeated with different individuals, varying lighting conditions and backgrounds, which demonstrates the practicability and robustness of our safe perception-based planning framework. In some situations, parts of the hand are detected by only a subset of the semantic segmentation models, showing the particularly strong impact of ensembling on safety.\looseness=-1
\section{Conclusion}
In this letter, we present a framework for vision-based motion planning with uncertainty using semantic image segmentation.
We show that combining 
massive data augmentation and deep ensembles yields good uncertainty quantification for semantic segmentation even for highly specific tasks lacking representative training data.
This allows us to 
interpret the output of the semantic segmentation probabilistically and use it for motion planning with uncertainty. 
We avoid the conservatism of existing uncertainty-aware path planning approaches by employing a sampling-based method for collision checking that is based on scenario optimization. As a result, our planning method makes no assumptions about the obstacle geometry and can be applied in highly cluttered environments.
Our 
framework is evaluated in simulation and experiment with a robotic manipulator. 
Directions for future work include using smaller semantic segmentation models, extending our method to estimate obstacle dynamics and enabling adaptive online planning based on a scenario MPC approach, so that it can be safely employed in dynamic environments.

\bibliographystyle{IEEEtran}
\bibliography{ref}

\end{document}